\documentclass[letterpaper, 10 pt, conference]{ieeeconf}  

\IEEEoverridecommandlockouts                              

\usepackage{graphicx} 
\usepackage{subcaption}
\usepackage{mathtools}
\usepackage{siunitx}
\usepackage{booktabs}
\usepackage{xcolor}

\title{\LARGE \bf
Considering Human Behavior in Motion Planning for Smooth Human-Robot Collaboration in Close Proximity}

\author{Xuan Zhao, Jia Pan
\thanks{X. Zhao and J. Pan are with the Department of Mechanical and Biomedical Engineering,
City University of Hong Kong, Hong Kong. This work was supported by Hong Kong Innovation and Technology Fund ITS/457/17FP and HKSAR
 Research  Grants  Council  (RGC)  General  Research  Fund
 (GRF) CityU 21203216.
        {\tt\small e-mail: xuan.zhao@my.cityu.edu.hk; jiapan@cityu.edu.hk}}
}

\begin{document}

\maketitle
\thispagestyle{empty}
\pagestyle{empty}

\begin{abstract}
It is well-known that a deep understanding of co-workers' behavior and preference is important for collaboration effectiveness. In this work, we present a method to accomplish smooth human-robot collaboration in close proximity by taking into account the human's behavior while planning the robot's trajectory. In particular, we first use an occupancy map to summarize human's movement preference over time, and such prior information is then considered in an optimization-based motion planner via two cost items as introduced in\cite{Hayne2016}: 1)~avoidance of the workspace previously occupied by human, to eliminate the interruption and to increase the task success rate; 2)~tendency to keep a safe distance between the human and the robot to improve the safety. In the experiments, we compare the collaboration performance among planners using different combinations of human-aware cost items, including the avoidance factor, both the avoidance and safe distance factor, and a baseline where no human-related factors are considered. The trajectories generated are tested in both simulated and real-world environments, and the results show that our method can significantly increase the collaborative task success rates and is also human-friendly. Our experimental results also show that the cost functions need to be adjusted in a task specific manner to better reflect human's preference. 
\end{abstract}
\section{Introduction} 
Many manufacturing tasks such as assembly require a high level of dexterity and flexibility beyond the capability of the state-of-the-art autonomous robotics technique, and thus a human worker's involvement is indispensable. One promising way to improve the efficiency of such challenging tasks is the human-robot collaboration, where human workers focus on sub-tasks requiring high flexibility and tactile sensing, while the robot leverages its high-speed and accuracy to quickly accomplish repetitive sub-tasks. A smooth and effective human-robot collaboration requires the human and robot to share a limited workspace. Such close proximity constraint leads to the popularity of collaborative robots because the traditional industrial robots must be isolated with human workers due to their large weight and high-speed.

\begin{figure}[thb]
\begin{center}
\begin{subfigure}[b]{0.23\textwidth}
\includegraphics[{width=\textwidth}]{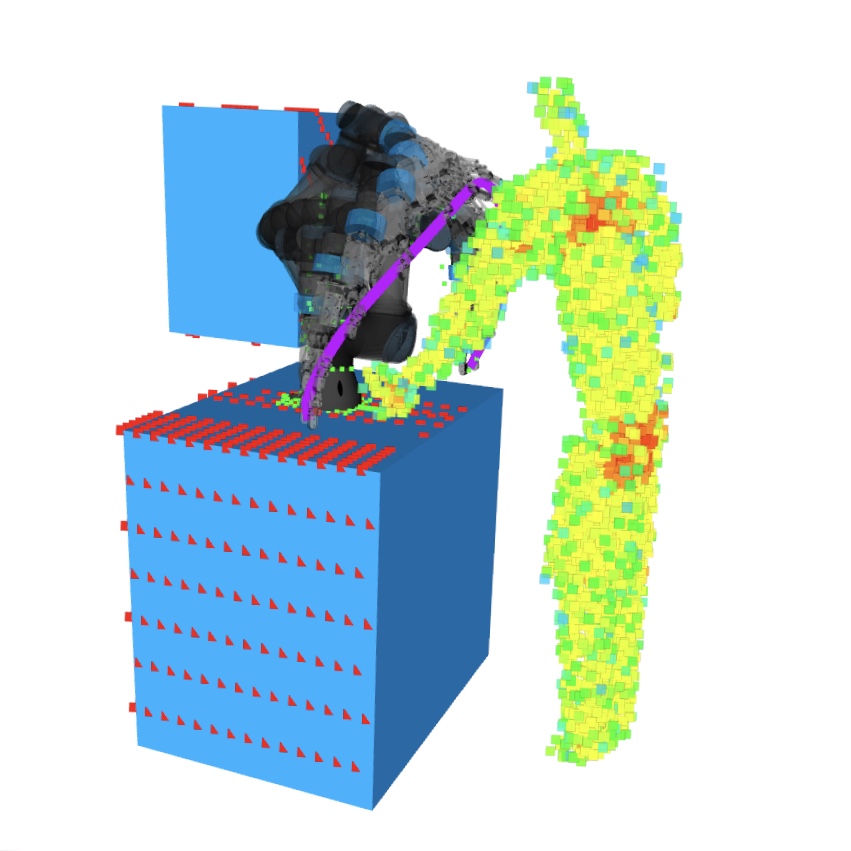}
\end{subfigure}
\begin{subfigure}[b]{0.23\textwidth}
\includegraphics[{width=\textwidth}]{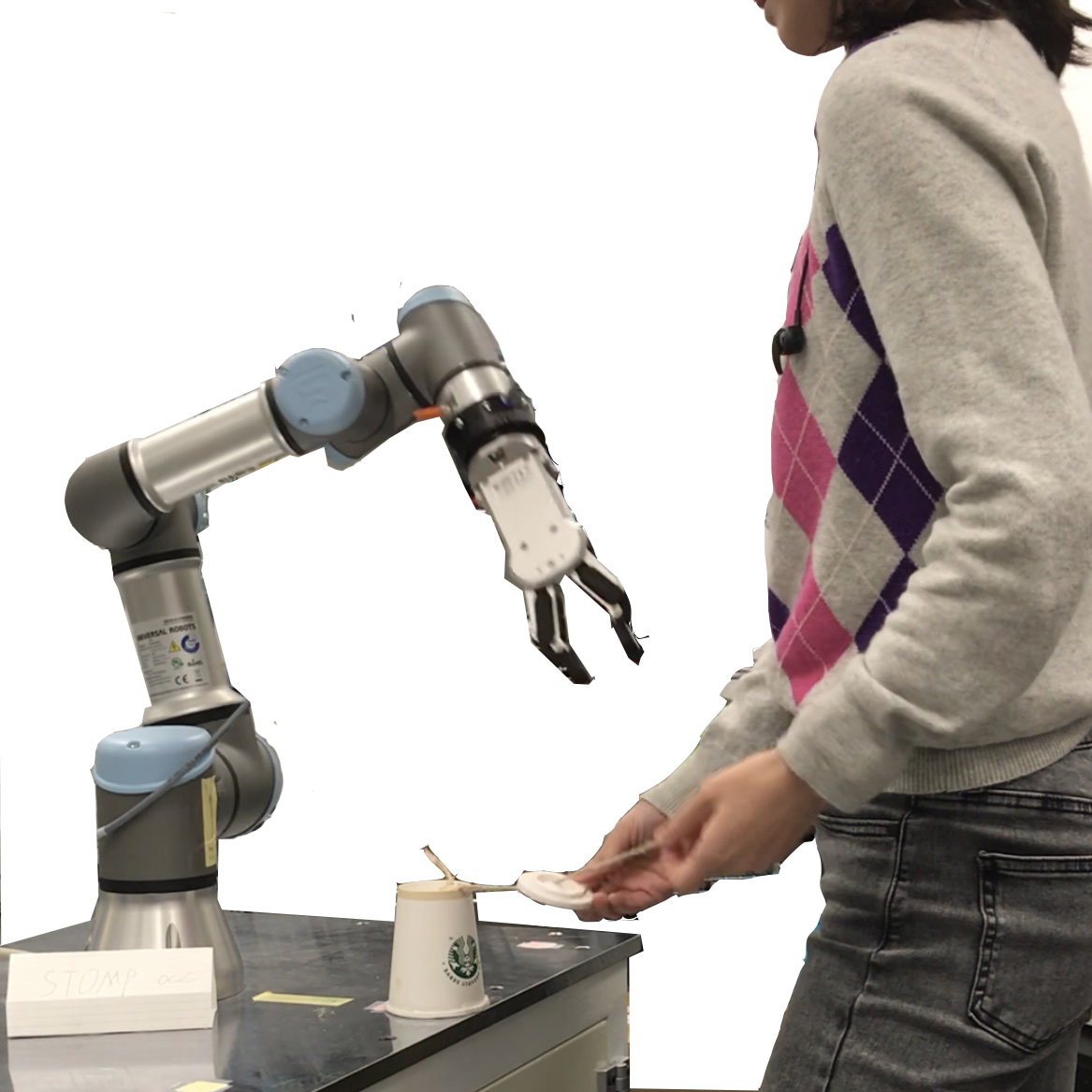}
\end{subfigure}
\caption{Close proximity human-robot collaboration by sharing a simulated environment (left) and a real-world scenario (right).}
\label{fig:sketch}
\end{center}
\end{figure}
\begin{figure*}[tb]
\begin{center}
\includegraphics[width=0.7\textwidth]{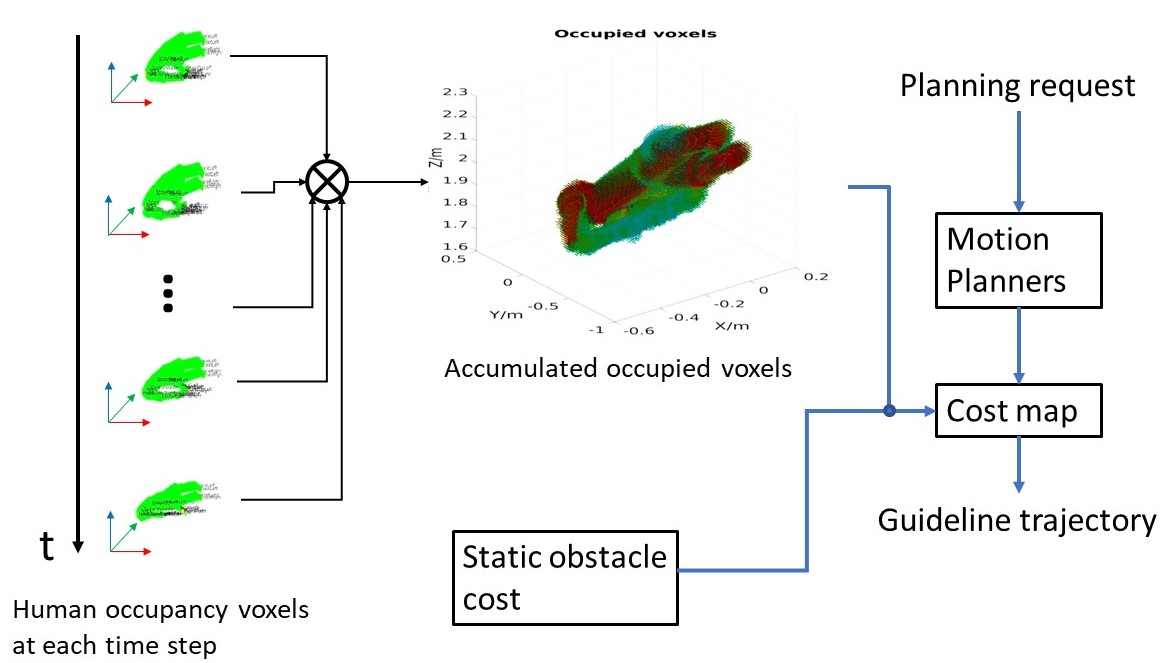}
\caption{The framework of our system. The pre-collected human behavior is used to generate the occupancy grid map. A cost map considering the avoidance and safe distance factors is generated from the occupancy grid. The static obstacles in the environment are also added to the cost map. This cost map and other optimization constraints are sent to the motion planner and to compute the trajectory.}
\label{fig:offlineframe}
\end{center}
\end{figure*}

To guarantee the safe collaboration with humans, the state-of-the-art collaborative robots work at a safe speed and must be turned into the emergency stop mode once a collision is detected. After the emergency stop, the robot needs to be restarted manually, which could significantly interrupt the workflow of both the robots and human co-workers. To resolve this issue, some recent human-robot collaboration approaches use online re-planning to actively avoid the perceived obstacles near the robot~\cite{mainprice2015predicting},~\cite{park2013real}. However, these collision avoidance methods do not take into account the prior information about the human's behavior, and thus the planned trajectory may not be consistent with the human's expectation, and thus may decrease the collaboration efficiency and increase the human's mental load. In addition, in close-proximity human-robot collaboration, the human motion is greatly constrained by the task and the surrounding environment and thus is more predictable than in the general situation. As a result, some methods pre-compute a set of candidate robotic trajectories and then switch online among different trajectories according to the current prediction of human motion~\cite{mainprice2013human},~\cite{luo2015framework}.

To make the robot aware of the human preference while planning its own motion, some recent methods accomplish smooth human-robot collaboration by using human-aware motion planners which use a cost map to explicitly encode a variety of human preference factors. For instance, for hand-over tasks,~\cite{mainprice2011planning} considers human's visibility, comfort, and reachability, while~\cite{sisbot2012human} considers handover positions, visibility, and force feedback. For manipulation tasks, collision avoidance and motion consistency factors have been used~\cite{Hayne2016}.

Similar to many previous works, our method also uses a cost map to encode the prior knowledge about human behavior.
In  Hayne \textit{et al.}'s ~\cite{Hayne2016} work, the authors considered two human-related factors: 1)~avoidance of the workspace frequently occupied by the human, and 2)~tendency to keep a safe distance from the human's high-frequency workspace. In our work, we use the same two two factors for our experiments about the assistive tasks where the robot needs to convey tools while the human co-worker is working in close proximity in order to reduce the interruption to the human. Compared to their manipulation tasks, experiments in our task show that humans prefer the robot motion generated from the planner only considering the first factor. This illustrates that for different tasks, the cost factors need to be adjusted accordingly to better reflect the human preference.

Our method generates the trajectory in an offline manner, but the generated trajectory can be used as a high-quality initial solution to the online planning framework with a low possibility for expensive re-planning. Our developed framework has three main steps as shown in Figure~\ref{fig:offlineframe}. First, we pre-collect a set of human movement data for accomplishing a task and generate an occupancy grid map which measures the human's accumulated occupancy frequency of the entire workspace. Next, given this occupancy grid map, we create a cost map for the two cost factors mentioned above, which is further combined with static obstacles in the environment. Finally, we send this cost map and optimization constraints to a motion planner to compute the robot trajectories, which are then tested in both simulated and real-world environments as shown in Figure~\ref{fig:sketch} to demonstrate the benefit of our methods for effective human-robot collaboration.

This paper is organized as follows. We describe the human behavior-related cost items and their application in optimization-based motion planning in Section~\ref{sec:approach}. We describe the experimental setup and demonstrate our results in Section~\ref{sec:result}.

\section{Approach}
\label{sec:approach}
In order to generate a safe motion avoiding human co-workers, our method uses a voxel-based representation to model the entire workspace and the region occupied by the human. The resulting human occupancy grids are used to define two cost functions with respect to human behaviors. The first cost item is the human occupancy cost which pushes the robot away from the region frequently occupied by the human. The second cost item is a signed distance function, which reflects the psychological safe distance between the human and the robot. Given these two cost functions, a robotic trajectory is then optimized using the STOMP~\cite{kalakrishnan2011stomp} motion planner. Our method is implemented on a Universal Robotics arm with $6$ degrees of freedom (DOF).

\begin{figure*}[tbp]
\begin{center}
\begin{subfigure}[b]{0.25\textwidth}
\includegraphics[{width=\textwidth}]{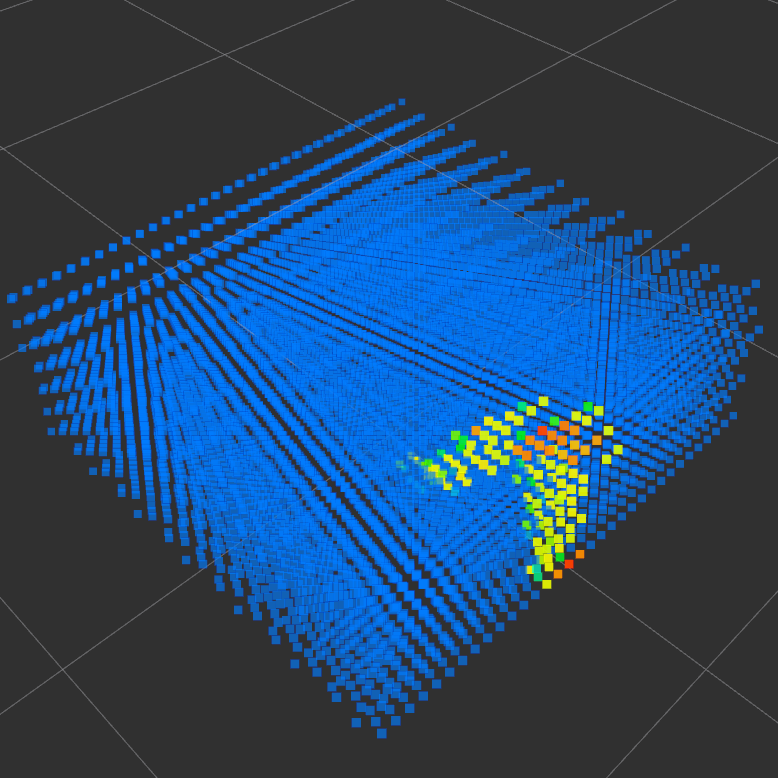}
\caption{A slice of the Occ map}
\label{fig:occh}
\end{subfigure}
\begin{subfigure}[b]{0.25\textwidth}
\includegraphics[{width=\textwidth}]{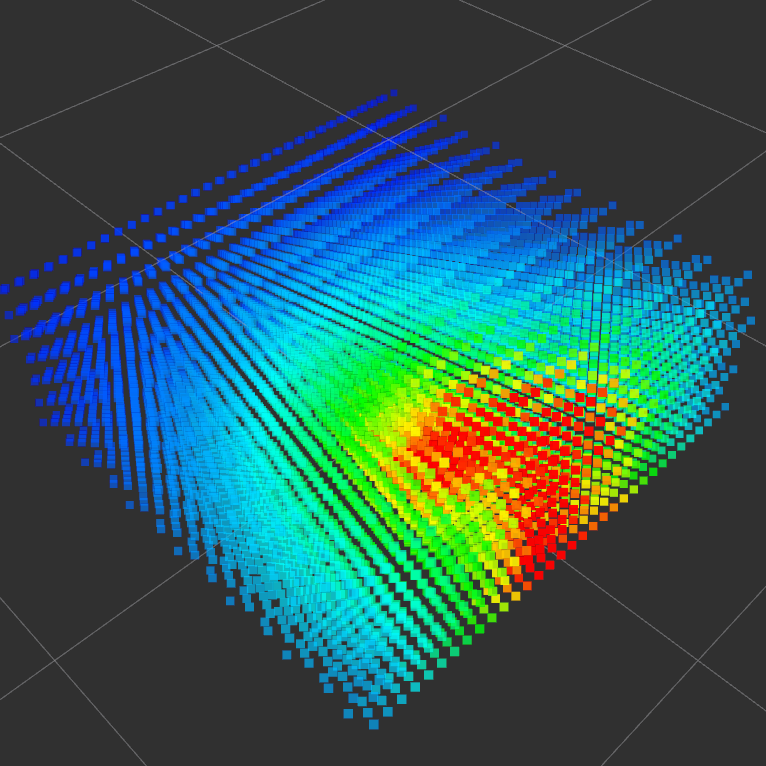}
\caption{A slice of the SDF map}
\label{fig:sdf}
\end{subfigure}
\begin{subfigure}[b]{0.25\textwidth}
\includegraphics[{width=\textwidth}]{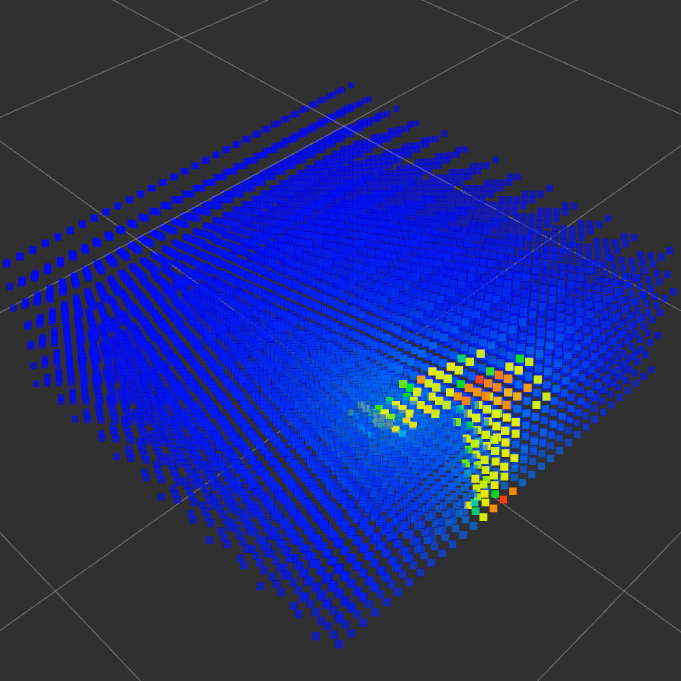}
\caption{A slice of the Occ$+$SDF map}
\label{fig:hpen}
\end{subfigure}
\caption{Cost maps generated from real human data. (a) shows the value of multi-layers of $occH(H,p)$, (b) shows the value of $sdfH(H,\mathbf{p})$, (c) shows the product of both maps, $pencost(H,\mathbf{p})$. Here the red color indicates large values and the blue color indicates small values.}
\label{fig:costmap}
\end{center}
\end{figure*}

\begin{figure*}[tbp]
\begin{center}
\begin{subfigure}[b]{0.25\textwidth}
\includegraphics[width=\textwidth]{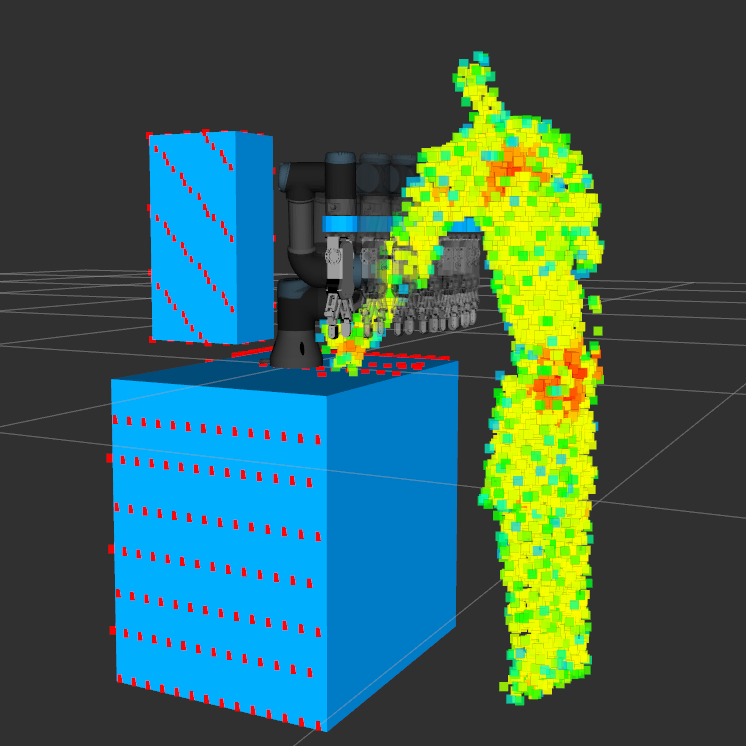}
\caption{RRT baseline method}
\label{fig:RRTbaseline}
\end{subfigure}
\begin{subfigure}[b]{0.25\textwidth}
\includegraphics[width=\textwidth]{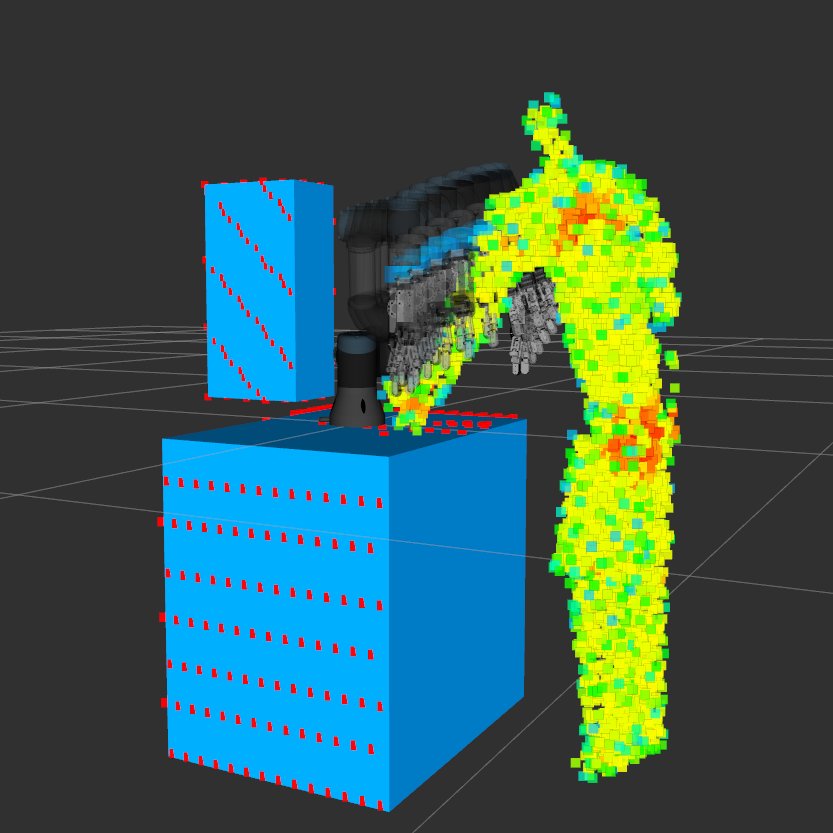}
\caption{STOMP with Occ map}
\label{fig:Stompocc}
\end{subfigure}
\begin{subfigure}[b]{0.25\textwidth}
\includegraphics[width=\textwidth]{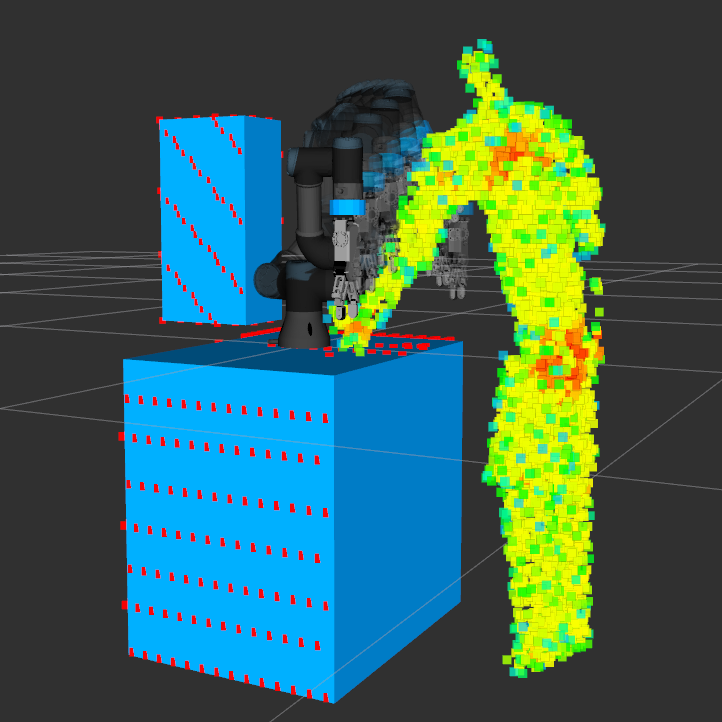}
\caption{STOMP with Occ$+$SDF map}
\label{fig:Stomphpen}
\end{subfigure}
\caption{Example of trajectories generated in a simulated environment by (a) the RRT baseline method, (b) the STOMP considering the occupancy and obstacle costs, and (c) the STOMP considering the occupancy, SDF and obstacle costs. Compared to the baseline method, both (b) and (c) reduce the percentage of penetration between robot trajectory and the region frequently occupied by the human.}
\end{center}
\end{figure*}

\subsection{Workspace Modeling}
The workspace is discretized into a set of voxel occupancy grids. We donate $H[i,j,k]$ as the human occupancy value each grid, where $[i,j,k]$ is the index of the voxel in the occupancy map. Each voxel $H[i,j,k]$ is initialized with an occupancy value of $0$.  A mapping function $m(\textbf{p}) = (i,j,k)$ maps a point $\textbf{p} = (x,y,z)$ to one grid index in the occupancy map. 

To update the human occupancy, we compute a set of sample points on the human body. The geometry of human body is modeled as a $17$-DOF link mechanism with $24$ cylindrical links. Each sample, donated as $\textbf{p}_{m}^{n}$, represents the $m^{th}$ point at the $n^{th}$ link. At each time step, a transform matrix $T_{n}^{0}$ maps the link from its original configuration to the current configuration. As a result, the human occupancy value in the grids can be updated as follows:
\begin{equation}
H[m(T_n^0\textbf{p}_{m}^{n})] = H[m(T_n^0\textbf{p}_{m}^{n})] +1. 
\end{equation}

\subsection{Cost Map Generation}
A moving human will not occupy one grid forever, and the human may actively adapt his movement to a robot's movement when the robot is in a near-collision state. As a result, the grids frequently occupied by the human are considered as a soft and high-cost region rather than impenetrable obstacles. In this way, a cost map is computed from the human's occupancy value in the workspace, which is then used by STOMP to guide the robot avoiding the regions frequently used by the human co-worker when achieving its own task.

In Hayne \textit{et al.}'s work~\cite{Hayne2016}, the authors formulated a cost map considering two factors related to the human behaviors. One is the avoidance factor as the occupancy (Occ) cost map, and the other is to push the robot away from the boundary of the region occupied by the human.
The Occ cost map is defined as:
\begin{equation}
occH(H,\textbf{p})=\left\{\begin{matrix} \frac{log(0.9+1)}{log(maxH+1))}, & \textrm{if} \ H[m(p)]=0 \\ \frac{log(H[m(p)+1])}{log(maxH+1)}, & \textrm{if} \ H[m(p)] > 0 ,\end{matrix}\right.
\end{equation}
where $maxH$ is the maximum voxel value across the entire human occupancy map. The log function is used to reduce the influence of the maximum voxel value because the linear normalization of raw $H$ voxel value will make the robot too sensitive to the maximum voxel value. For a grid outside the human's lane, we use the value $0.9$ while computing its occupancy cost rather the value $0$, in order to avoid the flat region in the occupancy map with zero cost. This setup also assigns a smaller cost to a non-occupied grid than a grid only occupied once by the human because the value $0.9$ is smaller than $1$.

In order to push the robot away from the boundary of the region occupied by the human, we also introduce the signed distance field (SDF) map, which is defined as:
\begin{equation}
sdfH(H,\mathbf{p}) = \frac{\arctan(maxSH)-\arctan(sdf(H,\mathbf{p}))}{\arctan(maxSH)-\arctan(minSH)},
\end{equation}
where $sdf(H, \mathbf{p})$ is the function that returns the SDF value for a given occupancy grid and a given query point $\mathbf{p}$. It will return a negative value when $\textbf{p}$ is inside the occupied volume described by the grid, a positive value if $\textbf{p}$ is outside the occupied region, and $0$ if $\textbf{p}$ is on the boundary of the occupied region. The magnitude is determined by how far $\textbf{p}$ is from the boundary. The $\arctan$ function is used for normalization because we want to reduce the influence points that are far away. $maxSH$ and $minSH$ are the maximum and minimum SDF value for the human occupancy map $H$. 
The occupancy map and the SDF map can be combined as follows:
\begin{equation}
pencost(H,\textbf{p}) = occH(H,\textbf{p})sdfH(H,\textbf{p}).
\end{equation}

Figure~\ref{fig:occh} and Figure~\ref{fig:sdf} show several layers of $occH(H, \mathbf{p})$ and $sdfH(H,\mathbf{p})$ respectively, and Figure~\ref{fig:hpen} shows the combination map $pencost(H,\mathbf{p})$ which indicates the penetration cost for the human.

There are also static obstacles in the workspace that need to be avoided completely. To avoid the optimization-based motion planner converges to a trajectory colliding with the static obstacles, the regions occupied by the obstacles are assigned with the highest cost. In particular, the obstacle cost is defined as follows:
\begin{equation}
obstacle(\mathbf{p})=\left\{\begin{matrix}1,&\mathrm{if}\  \mathbf{p}\ is\ in\ obstacle
\\0,& \mathrm{if}\ \mathbf{p}\ is\ not\ in\ obstacle.
\end{matrix}\right.
\end{equation}
By adding the human behavior cost map and the obstacle cost map together, we can get the final cost map: 
\begin{equation}
costmap(H,\textbf{p})=occH(H,\textbf{p}) + obstacle(H,\textbf{p}),
\end{equation}
or
\begin{equation}
costmap(H,\textbf{p})=pencost(H,\textbf{p}) + obstacle(H,\textbf{p}).
\end{equation}
\subsection{Robot Trajectory Optimization}
An optimization-based motion planner STOMP is used to compute an optimal trajectory~\cite{kalakrishnan2011stomp}. It minimizes the cost of a trajectory with respect to a variety of constraints. 
In our work, constraints include the collision avoidance, the final end-effector pose, the maximum joint velocity, and the maximum joint acceleration.

The trajectory cost includes two parts: the control cost and the state cost. The control cost is determined by the acceleration. The state cost of a trajectory is the sum of the state cost at each waypoint on the trajectory. The state cost for a waypoint is the sum of the cost values of the voxels occupied by the robot's current configuration. We donate $\mathbf{p}_{mr}^{nr}$ as the ${mr}^{th}$ point at robot link ${nr}$, and a mapping function $m(\mathbf{p})=(i,j,k)$. A transformation matrix $Tr_0^{nr}(s)$ represents the transformation between a link ${nr}$'s original configuration and its current configuration. The state cost at configuration $s$ is then defined as: 
\begin{equation}
cost(s)=\sum_{{nr}=1}^{N_r} \sum_{{mr}=1}^{M_{nr}}costmap(H,Tr_0^{nr}(s)\mathbf{p_{mr}^{nr}}),
\end{equation}
where $M_{nr}$ is the maximum number of sampled points on the geometry of link $nr$, $N_r$ is the number of robot links. 

\section{Results}
\label{sec:result}
In this section, we compared the trajectories generated by three methods: an RRT baseline method, STOMP with Occ cost map, and STOMP with Occ$+$SDF cost map. The cost and the length of each trajectory are calculated in a simulated environment. In addition, we tested the trajectories in two real-world human-robot collaborative tasks, and then analyzed the execution time, success rates, and the human preference. $8$ subjects, $6$ male and $2$ female, have participated in this experiment. Results show that our method is able to increase the task success rates, with both settings of the cost functions and the trajectories generated from the Occ map get the highest human preference score.

\begin{figure}[ht]
\begin{center}
\begin{subfigure}{0.22\textwidth}
\includegraphics[{width=\textwidth}]{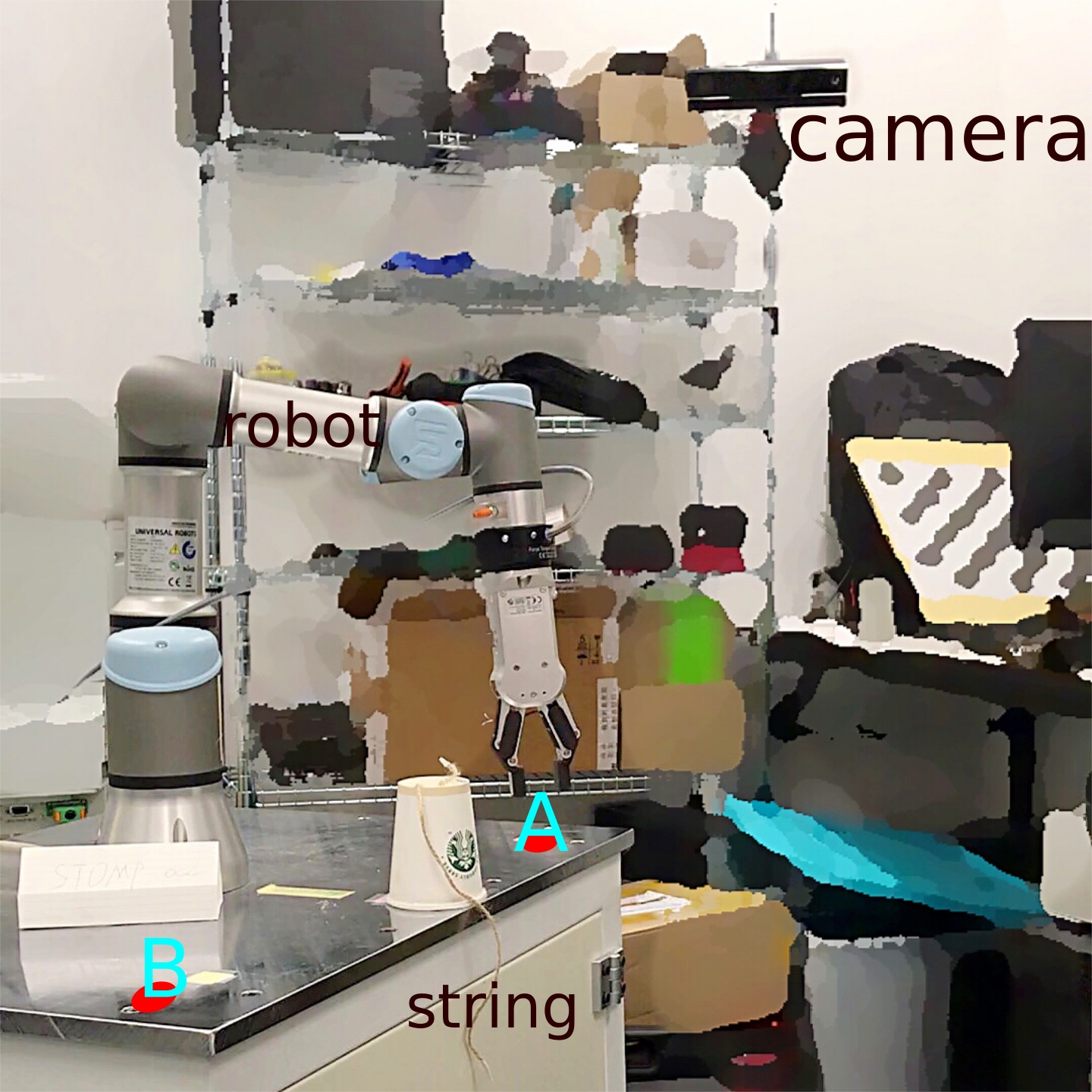}
\caption{}
\end{subfigure}
\begin{subfigure}{0.22\textwidth}
\includegraphics[{width=\textwidth}]{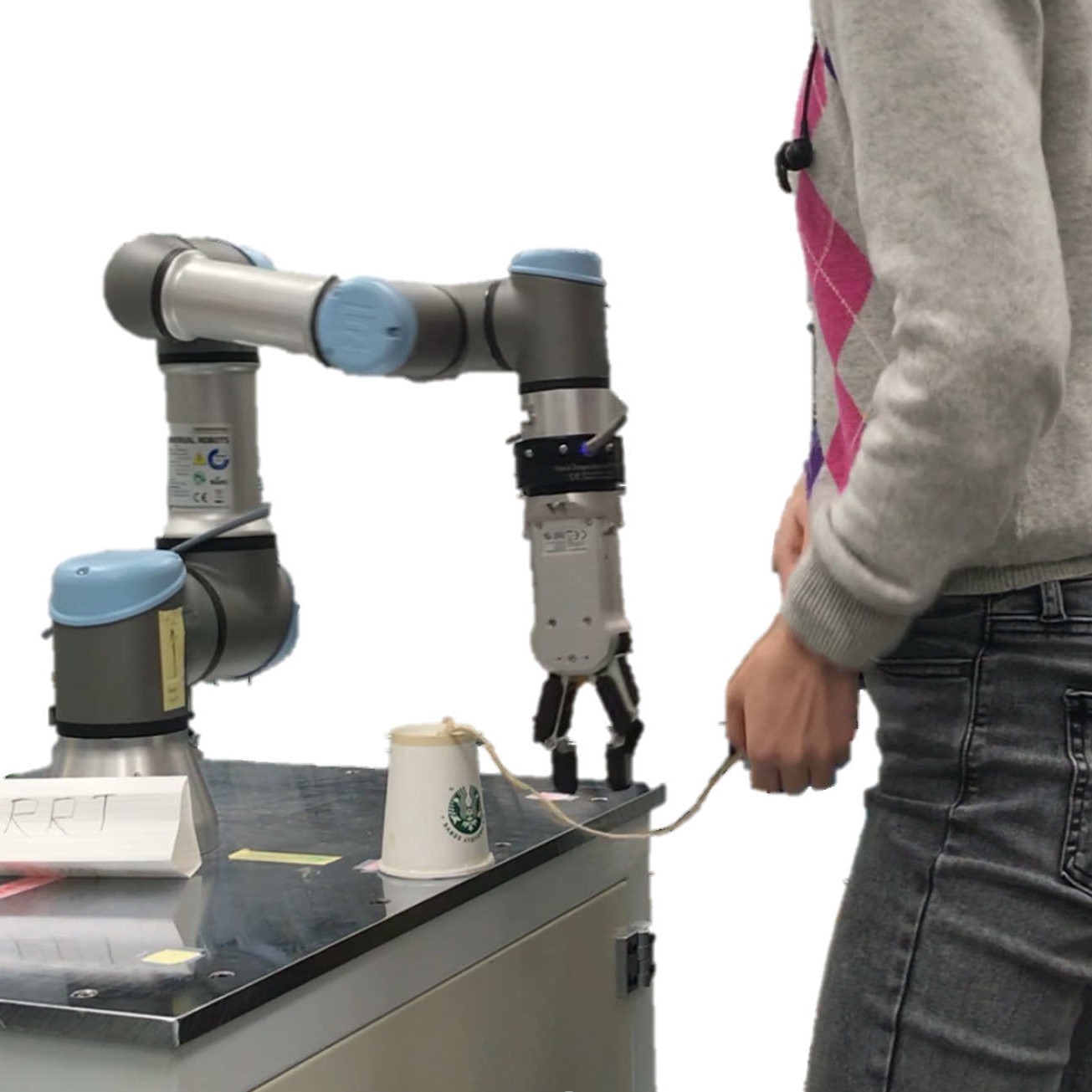}
\caption{}
\end{subfigure}
\begin{subfigure}{0.22\textwidth}
\includegraphics[{width=\textwidth}]{stomp_occ_real_2.jpg}
\caption{}
\end{subfigure}
\begin{subfigure}{0.22\textwidth}
\includegraphics[{width=\textwidth}]{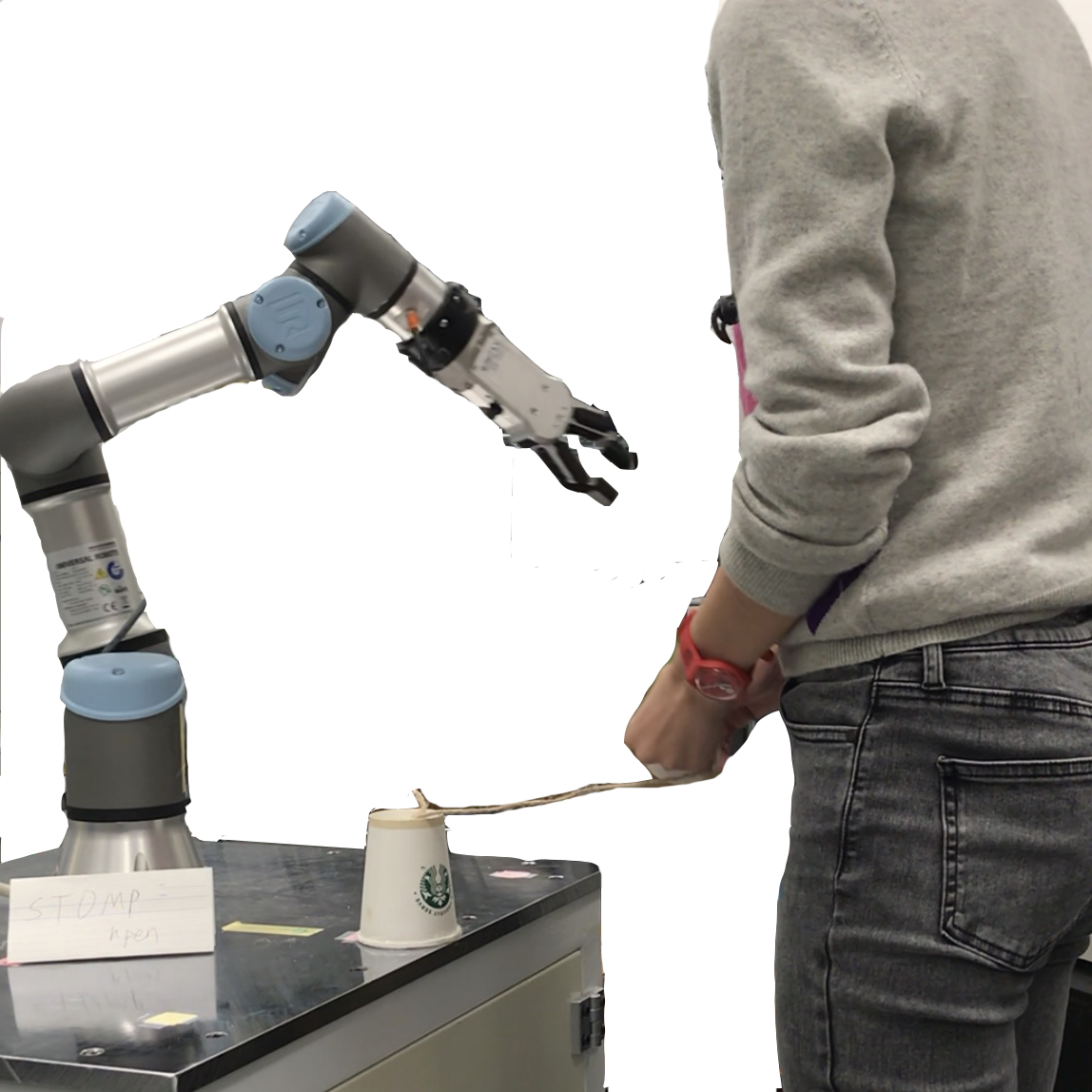}
\caption{}
\end{subfigure}
\caption{(a) The experimental setup. A robot needs to move from position A to position B, while human needs to perform a threading task with a long (20\si{cm}) or a short (5\si{cm}) string. (b) The RRT baseline trajectory directly moves through the human's working region and will interrupt human's work. (c) STOMP with only Occ map leaves the human's working area clean. (d) STOMP with Occ$+$SDF map also leaves the working area clean but generates a longer trajectory.}
\label{fig:expsetup}
\end{center}
\end{figure}
\subsection{Experiment Design}
The experimental setup is shown in Figure~\ref{fig:expsetup}. In this setup, a short (5\si{cm}) string and a long (20\si{cm}) string are fixed on the work desk. A human operator needs to manipulate the string to go through a small component, which is a deformable object manipulation task that is difficult for a robot. At the same time, the robot needs to move an object from position A to position B. If the robot blocks the human worker, the task will be counted as fail since the threading is a continuous task in which interruption is not allowed. The threading task with the shorter string requires the human operator to work closely with the robot's base, which increases the difficulty of the human-robot collision avoidance. The threading task with the longer string requires a longer operation time and also a larger space to accomplish, which implies that the human operator is likely to be blocked by the robot. Each human subject is asked to repeat two tasks for 20 times, and the procedure is recorded for computing the region frequently occupied by the human operator.

\begin{figure*}[tbp]
\begin{center}
\begin{subfigure}[b]{0.4\textwidth}
\includegraphics[width=\textwidth]{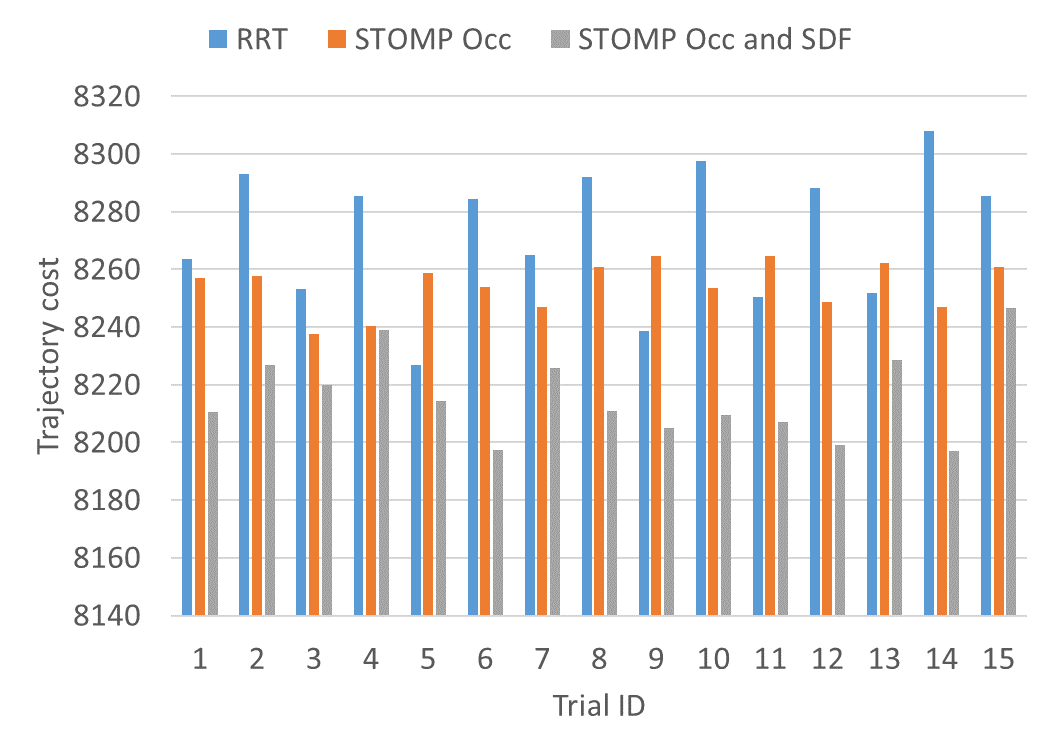}
\label{fig:trajectory_cost}
\end{subfigure}
\begin{subfigure}[b]{0.4\textwidth}
\includegraphics[width=\textwidth]{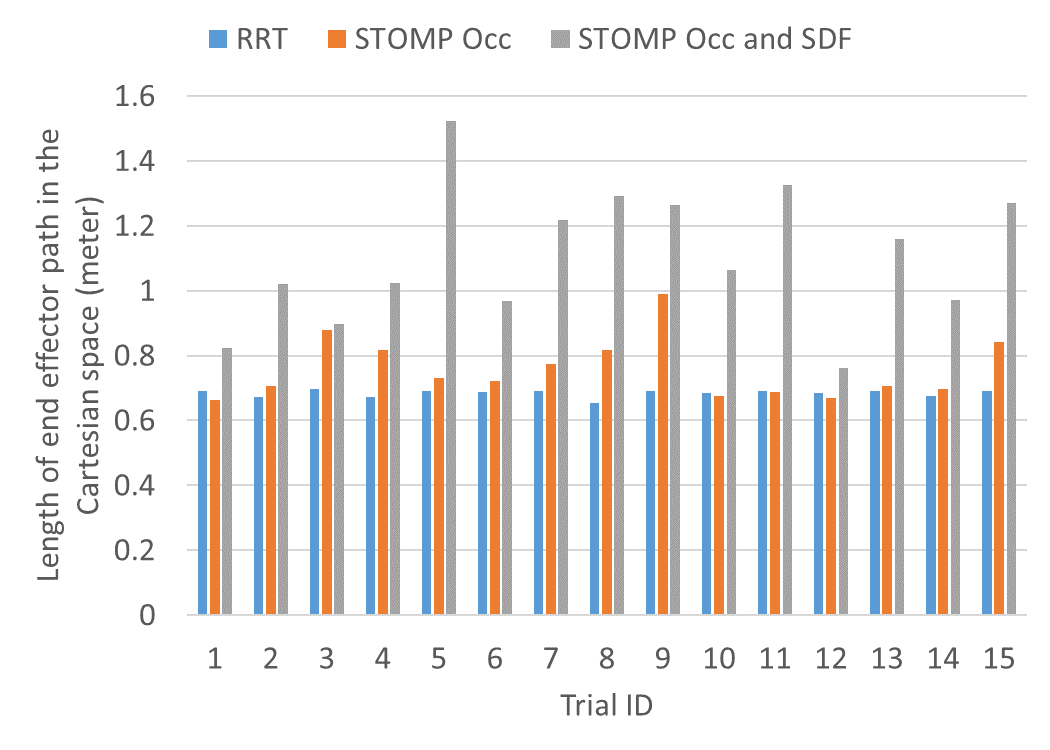}
\label{fig:trajectory_length}
\end{subfigure}
\caption{The trajectory cost in the Occ$+$SDF cost map (left) and the length of end-effector path in the Cartesian space (right). The baseline method has the highest cost and the shortest path, while the STOMP with Occ$+$SDF has the lowest cost and the longest path.}
\label{fig:traj_perform}
\end{center}
\end{figure*}

\begin{figure*}[tbp]
\begin{center}
\begin{subfigure}[b]{0.4\textwidth}
\includegraphics[width=\textwidth]{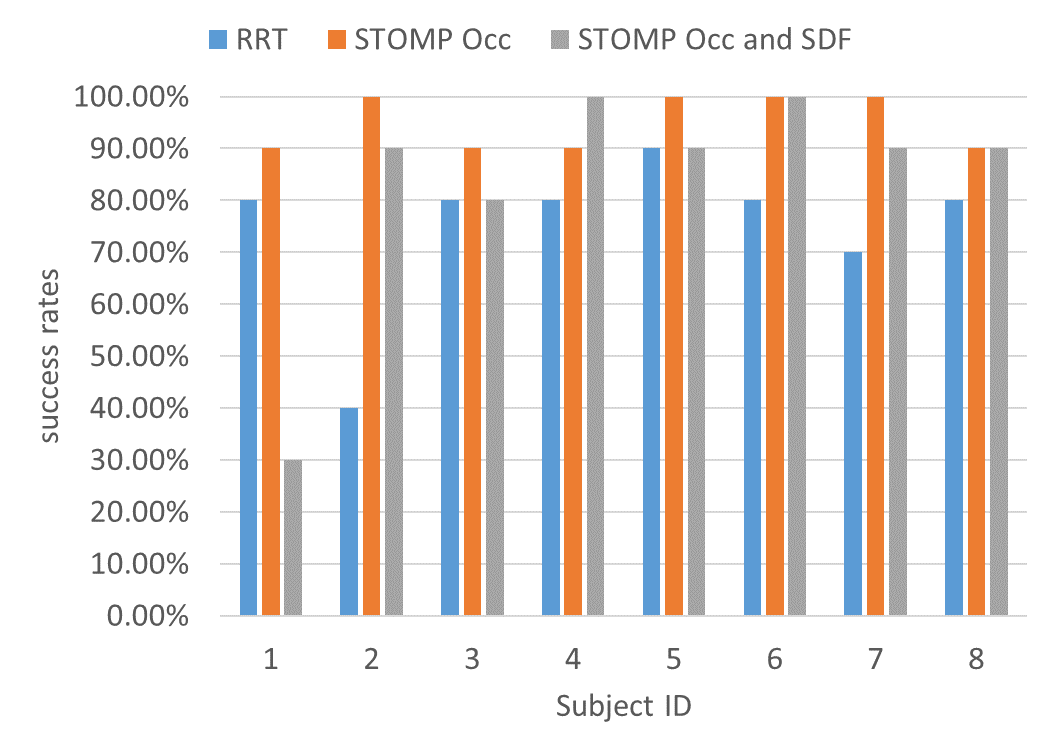}
\end{subfigure}
\begin{subfigure}[b]{0.4\textwidth}
\includegraphics[width=\textwidth]{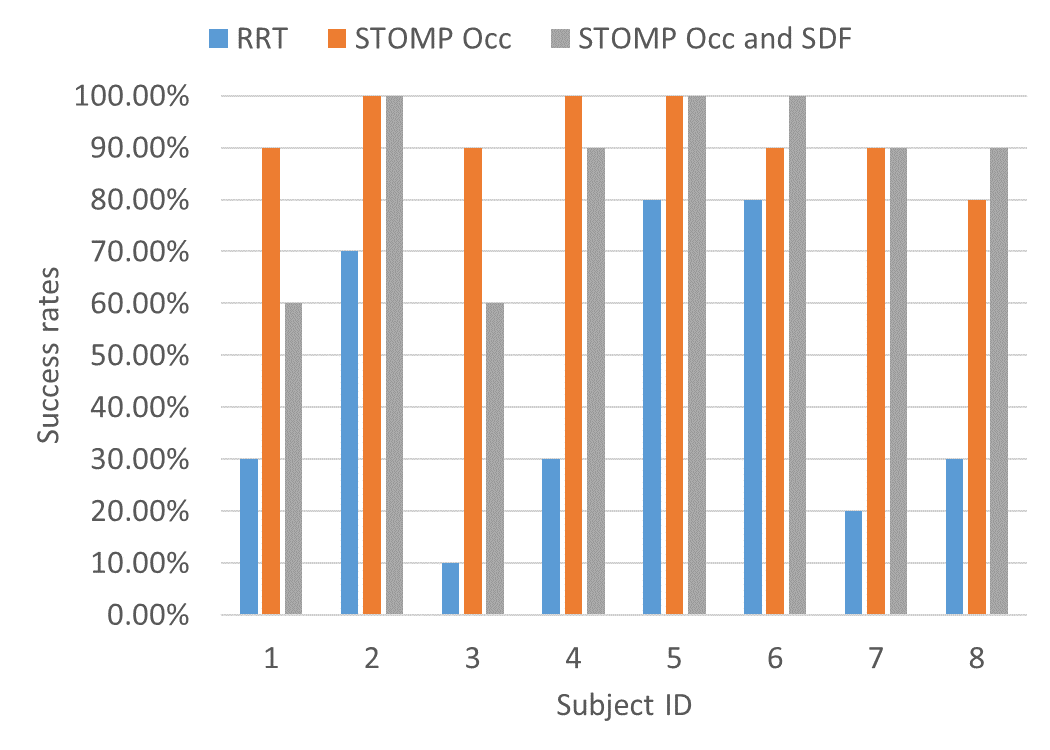}
\end{subfigure}
\caption{The success rates of the threading collaborative tasks involving the short string (left) and the long string (right). $8$ subjects participated in the experiment and each performed $10$ trials on both tasks with three methods. In this way, each subject performed a total of $3*2*10=60$ trials.} Results show that the task success rates are increased by taking into account human's behavior.
\label{fig:success_rate}
\end{center}
\end{figure*}

\subsection{Experiment Results}
Given the human's motion behavior, we then generate the robot's movement trajectory in an offline manner. Three different types of trajectories are computed, including a trajectory from a baseline RRT algorithm~\cite{lavalle1998rapidly} available in the OMPL library~\cite{sucan2012the-open-motion-planning-library}, one trajectory computed by the STOMP algorithm by only considering the Occ cost map, and one trajectory computed by STOMP by considering both the Occ$+$SDF cost map. Because both RRT and STOMP are stochastic planners, we ran $15$ trials for each method to compare their average performance. Example trajectories for each method in a simulated environment are shown in Figure~\ref{fig:RRTbaseline},~\ref{fig:Stompocc}, and~\ref{fig:Stomphpen} respectively.

We first compare three methods in the simulated environment where the recorded human occupancy map is given as the input. The evaluation criteria include the trajectory cost in the Occ$+$SDF map and the trajectory length. The trajectory cost is defined as:
\begin{equation}
trajcost(\Gamma) =  \sum_{i=0}^{W}cost(s_i),
\end{equation}
where $\Gamma$ is the robot trajectory in the configuration space. $s_i$ is the robot configuration at each waypoint $i$, and $W$ is the total number of waypoints of the trajectory. We are using $30$ waypoints evenly distributed along the trajectory according to the execution time. We compute the trajectory length cost as the length of the end-effector trajectory, which is defined as:
\begin{equation}
trajlen(\Gamma) =\sum_{i=1}^{W}\left \|  \boldsymbol{x}_e(s_i)-\boldsymbol{x}_e(s_{i-1})\right \|,
\end{equation}
where $\boldsymbol{x}_e(s_{i})$ is the position of the end-effector in the Cartesian space at the robot configuration $s_{i}$.

The results of simulated experiment are shown in Figure~\ref{fig:traj_perform}. The average trajectory cost of RRT, STOMP with Occ map, and STOMP with Occ$+$SDF map are: $8272.30\pm23.18$, $8254.23\pm8.21$, and $8215.83\pm14.47$ respectively.
Compared to the baseline method, both settings of our method reduce the trajectory cost with the lowest cost achieved by the STOMP considering the Occ$+$SDF cost map. Since the Occ$+$SDF cost map reflects the human occupancy and safe distance factors, lower cost indicates the lower percentage of penetration between the robot trajectory and the region frequently occupied by the human. The average trajectory length of three methods are: $0.68\pm0.01$, $0.76\pm0.09$ and $1.11\pm0.20\si{m}$ respectively. It shows that RRT gives the shortest but unsafe path, while STOMP with Occ$+$SDF gives the longest path. This is because the human's working region blocks the shortest path of the robot in the free space, and thus the robot needs to deviate and move longer distance to avoid the potential collisions and interruptions. In addition, STOMP with Occ$+$SDF cost map has a lower variance in the occupancy cost and the trajectory length compared to the STOMP with Occ cost map only.

\begin{table*}[htb]
\centering
\caption{Comparison of the average trajectory execution time, the task success rates for both tasks, and the human preference measurement based on the 1-5 Likert scale.}
\label{tab: human_preference}
\begin{tabular}{@{}l||llll@{}}
\hline
Method            & Trajectory execution time & Task success rates (short) & Task success rates (long) & Human preference (1-5 Likert scale) \\\hline \hline
Baseline          & $2.462s $                     & $75.00\pm14.14\%$                    & $43.75\pm26.43\%$                   & $1.125\pm0.331$            \\
STOMP Occ         & $3.083s$                     & $95.00\pm5.00\%$                    & $92.50\pm6.61\%$                   & $3.625\pm0.857$            \\
STOMP Occ and SDF & $3.969s$                     & $93.75\pm21.18\%$                    & $86.25\pm15.76\%$                   & $2.250\pm0.661$            \\ \hline
\end{tabular}
\end{table*}
Next, we tested the success rates of the collaborative tasks when human operators work with the robots in the real world scenario. Before the experiment, each subject was asked to read a description of the experiment and sign a consent form. $8$ subjects participated in this experiment and each performed $10$ trials for both the threading tasks involving the short string and the long string, using all the three methods. In this way, each subject performs a total of $3*2*10=60$ trials. To generate the robotic trajectory, we use three different initial trajectories to the STOMP since it is sensitive to the initial guess. One initial guess is the linear path connecting the start and the goal configuration, and the other two initial guesses are generated using the RRT planner. After the STOMP optimization, we choose the trajectory with the smallest cost as our final result. The trajectory is then executed by the robot in an open-loop manner to test its success rate while working with a human co-worker. Examples of a real-world collaboration test are shown in Figure~\ref{fig:expsetup}. As we can observe, the trajectory generated by RRT directly goes through the human's working region and interrupts the human's workflow. The trajectories generated by STOMP with both of our cost setups leave the human working region clean and rarely block the human's workflow.

The performance of the collaborative tasks is measured by three criteria, including the trajectory execution time, the task success rates, and the human preference. Table~\ref{tab: human_preference} demonstrates the result and Figure~\ref{fig:success_rate} shows the success rates of each subject. Compared to the baseline method, two methods involving the human behavior successfully increase the average task success rates. For each human operator, the same robotic trajectory is used for the 10 trials of each task, and we found that human subjects can quickly adapt to the robot motion and are able to find a proper way to finish both tasks after several trials even for the short string threading task which needs a high dexterity to be finished in a short time.

The human preference is measured by giving a 1-5 Likert scale to the subject after finishing all the trials. The statements on the chart are: "You prefer to work with the first/second/third trajectory." All the subjects gave the highest score to the second trajectory (generated from Occ map), and $7$ subjects gave the lowest score to the baseline trajectory. Subject 1 gave the lowest score to the third trajectory (which is calculated from Occ$+$SDF map). She responded that the third trajectory is too long to predict, which made her hesitate while working with the robot. However, subject 6 considered the long motion of the third trajectory as a useful notification of "I'm moving" from the robot. Most subjects replied that the third trajectory takes too long time to finish, so they need to wait. All the subjects agreed both the second and the third trajectories reduce the robot's invasion of the human operator's working region.   
\section{Conclusion}
\label{sec:conclusion}
In this work, we present a motion planning algorithm aware of the human motion behaviors to generate safe robotic trajectories for smooth human-robot collaboration in close proximity. The resulting trajectories are compared with a baseline planning method that does not take into account human co-workers, first in a simulated environment with the trajectory cost and length as the criteria and then in a real-world human-robot collaborative task with the execution time, task success rate and human preference as the criteria. Results show that our human-related cost functions can push the robot away from the human's working region. The resulting optimization-based planner has higher success rates of the collaborative tasks than the baseline method, and this is beneficial for the efficiency and smoothness of the collaboration. In addition, our method gets a higher score of human preference than the baseline method. 

Comparing to the previous work, there are also some interesting discoveries worth to notice. In Hayne \textit{et al.}'s manipulation tasks~\cite{Hayne2016}, human shows a higher preference when considering both Occ and SDF factors. However, in our experimental setup, there is a trade-off between the potential interruption and the trajectory execution time. Besides, adding the signed distance map to the Occ map does not significantly improve the task success rates but will result in a longer trajectory. These results a lower human preference when considering both Occ$+$SDF comparing to only Occ factors. Therefore, we conclude that for different tasks, to reach better human-friendly performance, the cost functions need to be adjusted according to the underlying tasks.

There are several limitations that we will resolve in future work. First, there are potential collisions because the trajectories are generated in an off-line manner. To overcome this issue, we will implement an online planning algorithm to keep the safe distance between a robot and a human. Second, we also observed that human workers have strong adaptability to the robot's behavior when the robot motion is fully consistent or predictable. Such fact will also be considered in future work. In addition, based on the users' feedbacks, especially the different ideas about the Occ$+$SDF trajectories, we want to further investigate other factors that may influence the human's psychological comfort for the motion planning.

\addtolength{\textheight}{-12cm}  
\bibliographystyle{IEEEtran}
\bibliography{IEEEabrv,bibliography}

\end{document}